\documentclass[conference,compsoc]{IEEEtran}
\ifCLASSOPTIONcompsoc
  \usepackage[nocompress]{cite}
\else
  \usepackage{cite}
\fi
\ifCLASSINFOpdf
\else
\fi

\usepackage{microtype}
\usepackage{graphicx}
\usepackage{subfigure}
\usepackage{booktabs} 
\usepackage{amssymb}

\usepackage{multirow} 
\usepackage{booktabs} 
\usepackage{times} 
\usepackage{graphicx} 
\usepackage{dblfloatfix} 
\usepackage[table]{xcolor}
\usepackage{float}
\usepackage{arydshln}

\usepackage{hyperref}

\newcommand{\comment}[1]{}

\usepackage{tikz}
\usepackage{textcomp}
\usepackage{hyperref}
\usepackage{lipsum}

\newcommand\copyrighttext{%
  \footnotesize \textcopyright 2021 IEEE SSCI. Personal use of this material is permitted.
  Permission from IEEE must be obtained for all other uses, in any current or future
  media, including reprinting/republishing this material for advertising or promotional
  purposes, creating new collective works, for resale or redistribution to servers or
  lists, or reuse of any copyrighted component of this work in other works.
  This is a preprint. The final version of the paper will be available in the IEEE SSCI 2021 Proceedings.}
\newcommand\copyrightnotice{%
\begin{tikzpicture}[remember picture,overlay]
\node[anchor=south,yshift=10pt] at (current page.south) {\fbox{\parbox{\dimexpr\textwidth-\fboxsep-\fboxrule\relax}{\copyrighttext}}};
\end{tikzpicture}%
}

\begin{document}
%
\title{Measuring Data Quality for Dataset Selection in Offline Reinforcement Learning}

\author{\IEEEauthorblockN{Phillip Swazinna}
\IEEEauthorblockA{Siemens Technology \& \\ Technical University of Munich\\
Otto-Hahn-Ring 6, Munich, Germany\\
Email: phillip.swazinna@siemens.com}
\and
\IEEEauthorblockN{Steffen Udluft}
\IEEEauthorblockA{Siemens Technology\\ \\
Otto-Hahn-Ring 6, Munich, Germany\\
Email: steffen.udluft@siemens.com}
\and
\IEEEauthorblockN{Thomas Runkler}
\IEEEauthorblockA{Siemens Technology \& \\ Technical University of Munich\\
Otto-Hahn-Ring 6, Munich, Germany\\
Email: thomas.runkler@siemens.com}}


%


\maketitle
\copyrightnotice
\begin{abstract}
Recently developed offline reinforcement learning algorithms have made it possible to learn policies directly from pre-collected datasets, giving rise to a new dilemma for practitioners: Since the performance the algorithms are able to deliver depends greatly on the dataset that is presented to them, practitioners need to pick the right dataset among the available ones. This problem has so far not been discussed in the corresponding literature. We discuss ideas how to select promising datasets and propose three very simple indicators: Estimated relative return improvement (ERI) and estimated action stochasticity (EAS), as well as a combination of the two (COI), and empirically show that despite their simplicity they can be very effectively used for dataset selection.
\end{abstract}


%
\IEEEpeerreviewmaketitle

\section{Introduction}
\label{intro}

Over the past years, reinforcement learning (RL) \cite{sutton1998introduction} algorithms have achieved outstanding performances in a variety of tasks and domains, such as robot locomotion, video games, and even industrial control \cite{mnih2013playing,lillicrap2015continuous,schulman2015trust,schulman2017proximal,haarnoja2018soft,depeweg2016learning,hein2016reinforcement}. Most algorithms are however only able to learn from data that is interactively collected during training, usually under or close to the current policy candidate, even though some are labeled "off-policy" algorithms. These approaches thus only work when granted interaction with the task environment in an "online" fashion and are generally unable to learn from a previously collected dataset \cite{fujimoto2018off}. Since perfectly accurate and unlimited use of simulations is generally unrealistic, this makes it hard to transfer the success that RL methods have had to more frequently encountered settings in real-world applications, where usually large datasets have been collected passively over time (e.g. autonomous driving, production facilities, turbines, traffic routing). In many cases practitioners do not have the opportunity to collect on-policy data, since doing so could be dangerous or at least prohibitively costly (e.g. turbine control, autonomous vehicles, reactors). Furthermore, it appears wasteful to not leverage large previously collected datasets to solve a problem, when they already exist in a lot of places.\\
To this end, more recently developed approaches have been specifically designed to work in this so-called "offline" reinforcement learning setting, where a new policy has to be derived from a single previously collected dataset. The resulting algorithms often feature elements of behavior cloning or so-called behavior regularization, so that new policies are not only optimized for performance, but also rewarded for closeness to the policy that generated the dataset, because that will in turn lead the chosen method of performance estimation to more likely be accurate \cite{fujimoto2018off,kumar2019stabilizing,swazinna2021overcoming,wu2019behavior,kumar2020conservative,swazinna2021behavior}. Other approaches use different ways of estimating the associated uncertainty to the policies' performance in order to regularize them, or directly try to train models under which the induced state-action distribution of the policy does not influence the prediction quality \cite{yu2020mopo,kidambi2020morel,yu2021combo,depeweg2016learning,kaiser2018bayesian}.\\
While the various new approaches to offline reinforcement learning have definitely led to improved applicability of reinforcement learning in practice, since the offline setting is very common in real-world problems, we argue that actually applying them is still problematic in most cases. For example, it has been shown that algorithms exhibit dramatic performance variances during training \cite{swazinna2021overcoming,fu2020datasets}, and since offline policy selection is still generally unsolved, users end up with the risk of deploying policies that are worse than simple behavior cloning. On top of that, many of the algorithms are rather hyperparameter sensitive, so they can perform well with tuned parameters, however it is not clear how to tune them offline, since offline policy estimation (OPE) methods are still struggling to perform consistently \cite{paine2020hyperparameter}. These issues could in practice be mitigated by choosing only very conservative hyperparameters (largely dishing the expected gains) or by allowing the evaluation of a few trained policies (where that is permissible).\\
Most crucially however, performances of offline RL methods vary drastically from dataset to dataset. This phenomenon is rather unsurprising since the methods can only learn from what has been demonstrated in the data that is presented to them. Companies wishing to benefit from the new offline RL technology thus face the issue of selecting which of their datasets are actually suitable to perform offline reinforcement learning on (alternatively whether such datasets even exist, or whether they would still have to be collected). Since the opportunity costs for such a project can be large (domain and ML experts time, compute resources, production environment stops for testing, possibly wasted / damaged materials, etc.), it would be interesting to know a priori whether the investment has a decent chance of payout. While benchmark datasets that have been published specifically for offline RL algorithms are usually all more or less suited, this is not necessarily the case in practice. In many companies, datasets will be available in a lot of places and more and more people will want to capitalize on them, so practitioners will have to decide which datasets have the best chances for success, and which others will likely only burn resources. In this paper, we will thus explore how practitioners could estimate the quality and suitability of their datasets for offline reinforcement learning, to make better use of limited resources.

\section{Related Work}
\label{related}
\textbf{Offline RL:} Batch Constrained Q-Learning (BCQ) \cite{fujimoto2018off} was one of the first Q-function based algorithms tackling the offline RL problem in continuous state and action spaces. It augments methods from behavior cloning, sampling likely actions from a model of the behavior policy, to constrain its policy to only select state-action pairs which the generating policy would have also likely chosen. Bootstrapping error accumulation reduction (BEAR) \cite{kumar2019stabilizing} picks up on this methodology, however embeds it in the actor-critic paradigm by optimizing a closed form policy constrained by the maximum mean discrepancy \cite{gretton2012kernel} between likely actions generated by the behavioral model and the action provided the policy. The behavior regularized actor critic (BRAC) \cite{wu2019behavior} framework generalizes the idea further and derives two new algorithms, BRAC-p (policy regularization) and BRAC-v (value function regularization), by additionally imposing a penalty on the trained Q-function. Conservative Q-Learning (CQL) \cite{kumar2020conservative} makes an effort to close the gap between estimated and real performance further, by deriving a value function whose expected value aims to lower bound the true performance of a policy. Many more model-free algorithms have been derived, such as CRR \cite{wang2020critic}, which filters samples based on their Q values, O-RAAC \cite{urpi2021risk}, which specifically optimizes for risk-averse decision making, and OPAL \cite{ajay2020opal}, which through hierarchical policies is able to optimize even long horizon tasks featuring sparse rewards from offline datasets. Other algorithms estimate policy performance with a transition model: MOOSE \cite{swazinna2021overcoming} directly transports the behavior regularization approach to a model-based setting, by simply switching out value function based performance estimation with a model-rollout approach, and WSBC \cite{swazinna2021behavior} uses recurrent transition models while regularizing the policies directly in weight space. MOPO, MOReL, MBPO, and COMBO \cite{yu2020mopo,kidambi2020morel,argenson2020model,yu2021combo} constitute hybrid approaches, that combine a transition model with value function based planning in a dyna fashion \cite{sutton1998introduction} in order to augment the initial dataset with more imagined data obtained through rollouts through the transition model.\\

\textbf{ML datasets:} In supervised machine learning, much progress has been made possible only by having large datasets such as e.g. Imagenet \cite{deng2009imagenet} accessible to researchers in order to better compare results. In an attempt to provide the same for offline RL, multiple works have proposed benchmark datasets for a variety of tasks and under various generating conditions. D4RL \cite{fu2020datasets} and RL unplugged \cite{gulcehre2020rl} aim to include undirected and multitask data, data collected by suboptimal agents, data generated from non-RL policies, as well as narrow data distributions on navigation, robotic locomotion, traffic control, as well as autonomous driving tasks. Neo RL \cite{qin2021neorl} tries to close the gap to reality further, by including datasets from more realistic problems (industrial control, electricity distribution, financial markets) and by generating the data in a more conservative, less exploratory fashion.\\

\textbf{Data quality:} In supervised learning, practitioners and researchers have already understood, that data quality is crucially important to a successful ML project. Even though most of academia focuses on improving algorithms, while earlier steps in the machine learning pipeline are getting less attention, methods addressing various data quality issues have been developed. Traditional approaches define data quality among the dimensions accuracy, completeness, consistency, and timeliness \cite{DQass,DQconsiderations}. Regarding accuracy, the concept of source trustworthiness \cite{federated} plays large role, and data quality can usually only be improved by collecting additional data: In \cite{ADLs}, the authors predict the same task labels with features collected from different sensors as data sources and find that even though the features reflect the same concepts, differences in data quality can have a drastic impact on classification performance. Similarly, \cite{water} finds that different camera sensors severely impact the quality of water turbidity estimation, and \cite{lidarcam} observe qualitative differences in depth estimations based on different sensors. A technique for improving accuracy in the absence of trustworthy sources is pointed out by \cite{noisylabel}, where many cheap labels from noisy labelers are combined.\\
Various data validation \cite{datavalidation, deequ, duckDQ, DaQL, TADQM} frameworks have been developed to address challenges relating to completeness, consistency, and timeliness: Deequ, DuckDQ, DaQL, and TFX enable practitioners to encode their expectations about how data may look like. Automated data cleaning methods based on anomaly or novelty detection offer the option to remove inconsistent data tuples even without any expert knowledge \cite{autoDQ, cleaning, NLPoutlier}.\\
In this paper we investigate how different data quality dimensions impact offline RL algorithms' performance on the respective datasets. Due to findings in the supervised setting as e.g. in \cite{ADLs, water, lidarcam}, it seems likely that differences in measured data quality will also play a significant role in RL. We are mainly concerned with two very simple data quality dimensions: (1) Quality of the best trajectories compared to the average trajectories, and (2) a notion of completeness that is rather close to the original quality definition in databases literature: Completeness is here referred to, as the "ability of an information system to represent every meaningful state of a real-world system" \cite{DQass, waw96}, which in RL basically boils down to how well datasets are explored and how accurate the resulting performance estimates will be.

\section{Problem Statement}
We wish to estimate data quality and in turn suitability of datasets for offline reinforcement learning. We assume to be given $N$ datasets of lengths $L_i$ consisting of tuples containing the state, action, reward, and future state of transitions $D_i = \{s_j, a_j, r_j, s_{j+1}\} \; i = 0..N-1, \; j = 0..L_i-1$. We are also given some metadata on the datasets and their respective tasks: Each dataset is associated with its own task-specific cost of deployment $C_i$ and all datasets incur a fixed cost $F$ when selected for development, reflecting expert time and compute cost. We are furthermore provided with domain knowledge about the ranges (maximum and minimum) of each state, action, and reward dimension. When a dataset is selected for development, offline RL algorithms are used to train a policy for deployment, and we observe a change in the average return of trajectories $\Delta R = \bar{R}^{\mathrm{data}} - \bar{R}^{\mathrm{new}}$. When this change is positive, it has the potential to offset our initial opportunity cost over the time horizon $H$, i.e. $R^{\mathrm{meta}} = \sum_{t=0}^H \left[\gamma^t\Delta R\right] - (C_i + F) \geq 0$. Ideally, we would like to select datasets with the largest $R^{\mathrm{meta}}$, without having to pay all $C_i + F$ (i.e. simply trying all out).

\section{Data Quality Estimation}
\label{methods_intro}
In the following, we will try to derive which dimensions in the data matter most to being suitable for using it to train a well performing policy using offline RL algorithms. Essentially, we identify two major settings which would be great scenarios for datasets to be used:
\begin{enumerate}
    \item Datasets which contain high return trajectories, while the average return is much lower. If trajectories or at least sub-trajectories with very high returns are present in the data, we would assume that the offline RL algorithms will be able to copy the behavior that has worked well, and find alternative solutions for where it has not.
    \item Datasets which contain high amounts of exploration. All considered algorithms feature either an action-value function or a dynamics model (or both) which is used to assess the performance of a policy and then provides a direction of improvement. When the dataset contains high amounts of exploration, i.e., large parts of the state-action space have been seen, we can realistically assume that the performance estimates become rather accurate, and thus the trained policies become better.
\end{enumerate}

We focus on dimensions and consequently also on deriving indicators that are broadly applicable and can thus be used for a variety of datasets that may be considered in practice. The indicators which we find are thus applicable in both low as well as high dimensional state spaces (including image data) and discrete as well as high dimensional action spaces.

\subsection{Return Expectation}
Identifying whether a dataset contains high reward trajectories is rather simple: We can just sum up the (discounted) logged rewards for each trajectory:
\begin{equation}
    \sum_{t=0}^{T} \gamma^tr_t
\end{equation}
For practitioners it can be useful to inspect the entire distribution of returns, also considering varying time horizons $T$. For simplicity we will however limit us to the maximum and mean of the observed full trajectory returns, which also appear to offer the most insight. Usually we would at least expect to outperform the mean observed return, since that would be easily attainable by simply employing behavior cloning methods. The more interesting anchor for expected policy performance will however likely be the maximum, since the algorithms should of course learn more from high return trajectories than from low return ones. Since different datasets from different domains may exhibit varying return scales, we need an indicator that is oblivious to the particular return scale. We thus define the estimated relative return improvement (ERI) as:
\begin{equation}
    \mathrm{ERI} = \frac{\max \left( R^{\mathrm{data}} \right) - \bar{R}^{\mathrm{data}}}{\bar{R}^{\mathrm{data}}} .
\end{equation}
For this ratio to be well defined, we need to assume that the returns are bounded from below by zero. If this is not the case (as seen later in the dataset analysis section), the returns need to be normalized by their minimum before further processing - as mentioned in the problem statement, we assume that such a minimum is simple domain knowledge and known or can be approximated reasonably well in practical applications.\\

Of course, offline RL algorithms can be expected to do more than simply copying good and avoiding bad trajectories. Rather, they should be able to intelligently combine good behavior on a more fine-grained level, however we will refrain from engineering indicators based on sub trajectory level returns in order to keep things as simple as possible.


\subsection{Estimating Exploratory Behavior}
After having analyzed the return distribution of the dataset, we move on to estimate the amount of exploration contained in it. Basically, we would like to know how much of the state-action space has already been seen, so that we can assess how confident we can be about performance estimates given by the value function or dynamics model. On a perfectly explored dataset, we would basically assume that the performance estimates can be estimated to a satisfactory degree, meaning that we would have a high chance of improving upon the baseline return.\\
We will start by analyzing the stochasticity of the generating policy. The generating policy is not assumed to be known, since in practice the RL practitioners usually have limited control over the collection process (datasets could have been collected over the past years of operation). Often, the behavior policy will thus be (close to) deterministic, but it does not have to be. It can also be a mixture of multiple policies, or even human interactions, both of which can be assumed to positively influence exploration. We simply train a $\mathrm{tanh}$ transformed gaussian policy, whose mean and standard deviation are given by neural network functions $f_{\mu}$ and $f_{\sigma}$, both operating on the same features $\phi(s)$, to match the state-action mapping in the dataset. $\theta$ denotes the joint set of parameters for $f_{\mu}$, $f_{\sigma}$, and $\phi(\cdot)$.
\begin{eqnarray}
    a \sim \mathrm{tanh}(\mathcal{N}(f_{\mu}(\phi(s)), f_{\sigma}(\phi(s))^2)) \\
    \theta^* = \mathrm{argmin}_{\theta} - \sum_t \log \; p(a_t|\theta, s_t) \label{eq:min}
\end{eqnarray}
The feature map $\phi(s)$ should be chosen to suit the state space of the problem. Since the datasets considered in section \ref{sec:experiments} all exhibit reasonably high dimensional state spaces, we chose to use a neural network with two hidden layers of size 100 and ReLU activations. For other spaces, other configurations may be better suited (e.g. adding convolutional layers for image spaces). After training the behavior policy by minimizing the negative log likelihood in Eq. \ref{eq:min} via gradient descent for 50 epochs, we iterate over the dataset to estimate the action stochasticity of the policy in each state, by inspecting the predicted standard deviations:
\begin{equation}
    \mathrm{AS}_f = \{f_{\sigma}(\phi(s))|s\in\mathcal{D}\}
\end{equation}
Practitioners can then analyze the distribution of the actions stochasticity $\mathrm{AS}_f$ to estimate exploratory behavior: Low values will indicate a rather deterministic and high values a very stochastic policy, which can be a decisive factor in dataset selection. For simplicity, we simply consider the mean estimated action stochasticity (EAS) in our later analysis. Note that this method has the tendency to overestimate stochasticity in regions of the state space that have not been visited (often) in the dataset.

\subsection{Other Relevant Data Quality Dimensions}
It could be worth knowing about other dimensions that may positively influence the chances of an offline RL algorithm to improve over the observed performance. \cite{qin2021neorl} claim that environment stochasticity can be a sufficient source of exploration, however we were unable to make this observation in our analysis, and we argue it also makes limited sense: If a deterministic policy sees many different trajectories and has thus 'explored' the state space, a subsequent learner has no way of knowing the influence of the taken actions, even if they may have led to a diverse set of successor states. \\

Hypothetically, it is possible that very little action stochasticity leads to very large changes in the observed trajectories, and that we thus underestimate the amount of exploration contained in a dataset if we only measure action stochasticity. An estimate of how much of the entire state space has been covered by a policy would thus be useful. Making a reasonable estimate is however extremely hard since (i) the problem becomes exponentially harder with increased dimensionality (ii) we may not even know where the extreme ends of our state space lie - the generating policy may have operated close to the boundary or very far from it. We experimented with a very basic indicator in this context: Selecting the minimum and maximum values encountered in each state dimension and calculate the volume of the resulting hypercube. Assuming expert knowledge about the true maxima and minima, we also calculate the hypercube volume of the entire problem and compute a ratio of the two. However, since we were unable to find any correlation with exploration or performance, we do not consider it further in this paper.

\begin{figure*}[h]
    \centering
    \includegraphics[width=1.2\textwidth]{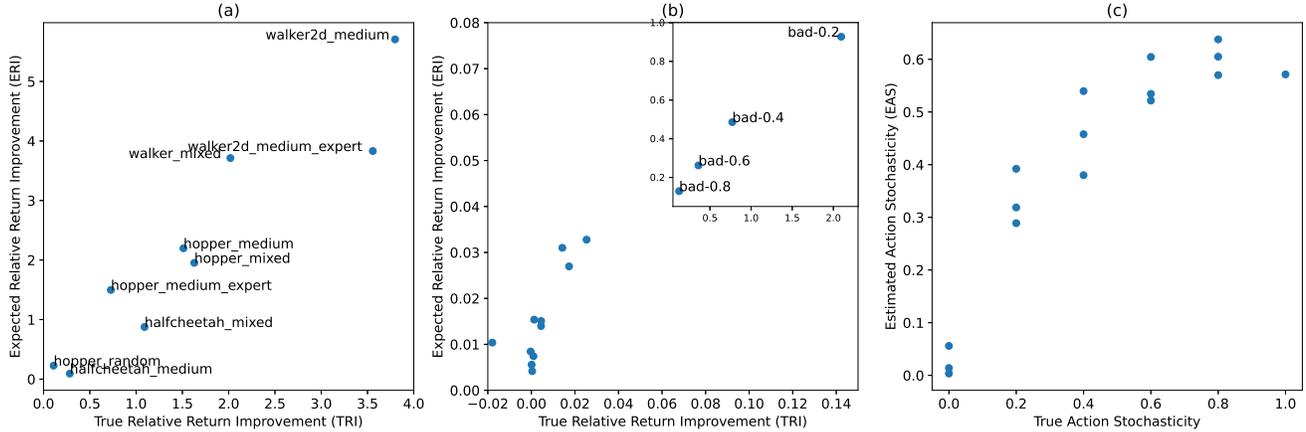}
    \caption{Data quality dimensions: Comparison of estimated and true values. Figure 1(a) shows the estimated relative return improvement (ERI) over the true relative return improvement (TRI) as given by the offline algorithm performance results in the MuJoCo datasets (note that halfcheetah\_random and walker\_random are extreme outliers and are not included in the graph). Figure 1(b) shows the same for the industrial benchmark datasets. Figure 1(c) shows the estimated action stochasticity (EAS) of the data generating policy together with the true action stochasticity in the industrial benchmark datasets - note that we cannot produce a similar graph for the MuJoCo datasets because we do not know exactly how stochastic the generating policies truly were, however we hypothesize that the result would look similarly well. The ranks of the indicators ERI and EAS as well as their correlation with TRI are presented in Figure \ref{table:ranks}.}
    \label{fig:graph_correlations}
\end{figure*}

\section{Analysis of widely used Offline RL Benchmark Datasets}
\label{sec:experiments}
We analyze 28 commonly used benchmark datasets for offline reinforcement learning algorithms taken from \cite{fu2020datasets} and \cite{swazinna2021overcoming}, which were collected using a variety of policies in different MuJoCo continuous control tasks and the industrial benchmark (IB) \cite{hein2017benchmark}. While multiple datasets originate from the same environment in reality, we assume that we do not have this information for the sake of our experiments, since otherwise in practice one would simply combine all available datasets together to have the most comprehensive knowledge about the problem to be solved. We thus treat each dataset as independent of the others. We use algorithm performance results that were published together with the datasets to determine a ground truth for their real improvement / suitability with which we can then compare our predictions about how well we estimate them to be suited. The ground truth relative improvement is calculated as:
\begin{equation}
    \Delta R = \frac{R^{\mathrm{algo}} - \bar{R}^{\mathrm{data}}}{\bar{R}^{\mathrm{data}}}
\end{equation}
where $R^{\mathrm{algo}}$ is the reported performance of the offline algorithm and $\bar{R}^{\mathrm{data}}$ is the average performance as observed in the dataset. As mentioned in Section \ref{methods_intro}, we need to assume that returns are bounded from below by zero in order for the ratio to be well defined. Since the (IB) datasets and also some of the MuJoCo datasets contain negative return trajectories, we normalize them with reasonable minima (-350 for the IB and the lowest observed return across the same task for MuJoCo). \\

Figure \ref{fig:graph_correlations} presents the data quality indicators ERI and EAS, that we calculated according to their definition as given in section \ref{methods_intro} together with the ground truth values for these dimensions. It is noteworthy how well both estimated relative return improvement (ERI) and estimated action stochasticity (EAS) coincide with the reality. Albeit not perfect, it seems that the quality dimensions can be well estimated. The ranks and rank correlations of the datasets with respect to both ERI and EAS are presented in Figure \ref{table:ranks}. 

The EAS indicator generally correlates with TRI weaker than the ERI does, however this makes sense since its impact is a little less direct: A lot of exploration may be very good for downstream RL algorithms, however whether it actually results in better performance depends also on other factors: (i) Even if a lot of exploration has been performed, that does not mean that the right areas of the state-action space have been seen. While halfcheetah\_random exhibits a TRI of almost 16, walker\_random only has a factor of about 4.5, and hopper\_random is only slightly above zero. (ii) In the case of the 'bad' baseline datasets on the IB, the exploration is actually inversely correlated to true relative improvement, since the baseline is so bad that random actions perform much better, leaving less room for improvement. As shown in Figure \ref{table:ranks}, if the 'bad' datasets were omitted, the rank correlation here would thus improve from $0.13$ to $0.83$. In reality, the impact of the exploration is however still likely to be lower than the ERI, since exploration will surely lead to return diversity and thus an increasing ERI, which is something we can also observe empirically.\\

To get a definitive ordering by which to select datasets, we thus combine ERI and EAS to a combined indicator COI (combined offline indicator) by weighing the ranks of ERI and EAS 2:1. Figure \ref{table:ranks} also includes the datasets' rankings by this new combined indicator, as well as by the true relative return improvement (TRI) and their rank correlation. Scoring between 0.75 and 0.86, COI correlates very well with the ground truth improvement TRI, meaning that our indicators are in fact suitable to detect good offline datasets where we have a high chance of improving upon the previously observed average return with offline RL.\\
\newpage
Figure \ref{table:ranks} also divides the datasets in a top and bottom half by their TRI and marks COI ranks in bold if the rank is in the correct half in terms of TRI. In a fictitious example where we would have had the ability to work on half of the projects that are proposed to our data science department, we would have selected almost precisely the upper half of the projects with the best true improvements, if we had selected the datasets based on COI. Other selections, such as the best or the top three datasets would have been less optimal. However, it is unreasonable to assume that simple data quality indicators could predict performance increases on an individual level for two reasons: (i) the dataset may be anomalous in the sense that either the action stochasticity was unhelpful (wrong part of the state-action space explored), or (ii) the algorithm used may for some reason fail to realize the potential of the dataset: An example here would be CQL on the halfcheetah\_medium\_expert dataset, where it achieves below the average dataset return, even though a performance increase of 50\% would have been possible simply by imitating the upper half of transitions. Our indicator becomes thus much more helpful, the more datasets are selected for work since individual outliers can be compensated. However, even in the case of only a single dataset being selected, our indicator would be in the upper half, avoiding at least worst-case decisions.

\begin{figure*}[t]
\parbox{.45\linewidth}{
\centering
\begin{tabular}{l|c|c|c||c|}
     Dataset & ERI & EAS & COI & TRI \\
     \hline
    bad-0.2 & \textbf{15} & 3 & \textbf{10} & 15\\
    bad-0.4 & \textbf{14} & 5 & \textbf{11} & 14\\
    bad-0.6 & \textbf{13} & \textbf{8} & \textbf{12} & 13\\
    bad-0.0 & \textbf{8} & 0 & 5 & 12\\
    bad-0.8 & \textbf{12} & \textbf{14} & \textbf{15} & 11\\
    mediocre-1.0 & \textbf{11} & \textbf{12} & \textbf{13} & 10\\
    mediocre-0.6 & \textbf{9} & \textbf{13} & \textbf{9} & 9\\
    mediocre-0.8 & \textbf{10} & \textbf{15} & \textbf{14} & 8\\
    \cdashline{2-5}
    optimized-0.8 & \textbf{6} & 11 & 8 & 7\\
    mediocre-0.4 & \textbf{5} & 10 & \textbf{7} & 6\\
    mediocre-0.2 & \textbf{7} & \textbf{6} & \textbf{6} & 5\\
    optimized-0.4 & \textbf{2} & \textbf{7} & \textbf{3} & 4\\
    optimized-0.0 & \textbf{0} & \textbf{1} & \textbf{0} & 3\\
    optimized-0.2 & \textbf{1} & \textbf{4} & \textbf{1} & 2\\
    optimized-0.6 & \textbf{3} & 9 & \textbf{4} & 1\\
    mediocre-0.0 & \textbf{4} & \textbf{2} & \textbf{2} & 0\\
     \hline
     Spearman's $\rho$ to TRI & 0.91 & 0.13 & 0.75 &  \\ 
     $\rho$ w/o 'bad-*' datasets & 0.81 & 0.83 & 0.86 &  \\
     \hline
     
\end{tabular}
}
\hfill
\parbox{.5\linewidth}{
\centering
\begin{tabular}{l|c|c|c||c|}
    Dataset & ERI & EAS & COI & TRI \\
     \hline
     halfcheetah\_random & 5 & \textbf{10} & \textbf{8} & 11\\
     walker\_random & 3 & \textbf{11} & \textbf{7} & 10\\
    walker\_medium & \textbf{11} & 3 & \textbf{9} & 9\\
    walker\_medium\_expert & \textbf{10} & \textbf{7} & \textbf{11} & 8\\
    walker\_mixed & \textbf{9} & \textbf{8} & \textbf{10} & 7\\
    hopper\_mixed & \textbf{7} & 2 &  5 & 6\\
    \cdashline{2-5} 
    hopper\_medium & 8 & \textbf{0} & \textbf{4} & 5\\
    halfcheetah\_mixed & \textbf{4} & 6 & \textbf{3} & 4\\
    hopper\_medium\_expert & 6 & \textbf{5} & 6 & 3\\
    halfcheetah\_medium & \textbf{0} & \textbf{4} &  \textbf{1} & 2\\
    hopper\_random & \textbf{1} & \textbf{1} &  \textbf{0} &  1\\
    halfcheetah\_medium\_expert & \textbf{2} & 9 & \textbf{2} & 0\\
    \hline
    Spearman's $\rho$ to TRI & 0.59 & 0.39 & 0.85 &  \\
    \hline
\end{tabular}
}
\caption{Datasets sorted by the rank of their true relative return improvement (TRI), where higher is better. We show the ranking of the datasets with regard to the derived indicators estimated relative return improvement (ERI), estimated action stochasticity (EAS), and the 2:1 weighted combination of their ranks (COI = combined offline indicator). The dashed line separates the datasets in a top and bottom half based on the true relative return improvement, and bold ranks of COI mark the datasets that would have been correctly identified into the respective half. In a setting where half of the datasets should be selected for work, the selection based on COI is thus almost entirely correct.}
\label{table:ranks}
\end{figure*}

\section{Conclusion}
In this paper we presented a new perspective on offline reinforcement learning: The recently popularized approach has brought RL much closer to practical applicability, however most current literature is still mainly concerned with deriving algorithms to make learning from datasets possible in general, while we propose a new perspective on the problem in practice: Often, there exist many previously collected datasets for industrial control problems instead of just one. Hence, suitable datasets need to be selected among the plethora of available ones as there are usually not enough resources for practitioners to work on all projects at once. We thus formulate the \textit{dataset selection problem} and discuss which dimensions of data quality are important factors in an offline RL algorithm's ability to improve upon the mean performance in a dataset. We find mainly two valuable data quality indicators: Estimated relative return improvement (ERI) and estimated action stochasticity (EAS). We analyze the indicators in terms of their effect on algorithm performances on common benchmark datasets and propose a final combined indicator, which constitutes a robust approach to the \textit{dataset selection problem}.\\

Our work is guided by the assumption that methods exist to examine a dataset for its approximate worth in offline reinforcement learning, which should be much simpler and less expensive than simply trying out algorithms and evaluating their performance on real systems. It is clear that the proposed indicators are far from perfect, however we argue that this is neither possible nor necessary. It is not possible since offline RL algorithms may be arbitrarily complex, hyperparameter sensitive, or stochastic, and may fail to realize a datasets theoretic potential. However, perfection is luckily not necessary to improve decision making on average in the long run: Any indicator that has a positive correlation with the true improvement can achieve this. Interestingly, we find that very simple indicators already correlate strongly with the true improvement, suggesting that algorithms are not as complex as expected, and that their performance can be better explained with simple indicators than one may expect. Since the two indicators ERI and EAS reflect the two most commonly used (and arguably most important) sources of improvement - bias towards high return trajectories and exploration - we find that it makes sense not only empirically, but also from a theoretical perspective to use them for dataset selection in practice.\\

Our contributions are to point out and discuss a new problem setting that arises due to offline reinforcement learning that has so far not been considered and an initial approach for a robust solution, based on three simple, yet efficient indicators for dataset quality that can be helpful for dataset selection. We evaluate the proposed indicators and show that ERI, EAS, and especially COI constitute a robust approach to the \textit{dataset selection problem}.

\section{Future Work}
Future work in this area may include the search for different indicators that influence algorithm performance to further improve dataset selection. Especially our findings regarding state-space coverage were unsatisfying in this regard. Another direction of research could be the anlysis of algorithm anomalies found by the proposed approach, and in turn improving algorithms: As mentioned earlier, CQL performed below average on 'halfcheetah\_medium\_expert', even though a 50\% improvement would have been easily achievable. Furthermore, measuring certain data quality dimensions such as exploration could be used to tune common hyperparameters in offline RL algorithms that control how close the new policy needs to stay to the behavior policy.

\ifCLASSOPTIONcompsoc
  \section*{Acknowledgments}
\else
  \section*{Acknowledgment}
\fi
The project this paper is based on was supported with funds from the German Federal Ministry of Education and Research under project number 01\,IS\,18049\,A.



%



\end{document}